\documentclass[letterpaper]{article} % DO NOT CHANGE THIS
\usepackage{aaai23}  % DO NOT CHANGE THIS
\usepackage{times}  % DO NOT CHANGE THIS
\usepackage{helvet}  % DO NOT CHANGE THIS
\usepackage{courier}  % DO NOT CHANGE THIS
\usepackage[hyphens]{url}  % DO NOT CHANGE THIS
\usepackage{graphicx} % DO NOT CHANGE THIS
\usepackage{natbib}  % DO NOT CHANGE THIS AND DO NOT ADD ANY OPTIONS TO IT
\usepackage{caption} % DO NOT CHANGE THIS AND DO NOT ADD ANY OPTIONS TO IT
\usepackage{algorithm}
\usepackage{algorithmic}

\usepackage{newfloat}
\usepackage{listings}
\usepackage{caption}
\usepackage{graphicx}
\usepackage{float} 
\usepackage{subfigure}
\usepackage{amsmath}
\usepackage{mathrsfs}
\usepackage{amsmath}
\usepackage{amsthm}
\usepackage{booktabs}
\usepackage{algorithm}
\usepackage{algorithmic}
\usepackage{multirow}
\usepackage{amssymb}
\usepackage{bbding}
\usepackage{mathrsfs}
\usepackage[normalem]{ulem}
\usepackage{color}
\usepackage[dvipsnames]{xcolor}
\usepackage{microtype}
\usepackage{amsmath}
\usepackage{amsthm}
\usepackage{booktabs}
\usepackage{algorithm}
\usepackage{algorithmic}
\usepackage{multirow}
\usepackage{amssymb}
\usepackage{bbding}
\usepackage{mathrsfs}
\usepackage{longtable}
\usepackage[T1]{fontenc}
\usepackage{enumitem}
\DeclareCaptionStyle{ruled}{labelfont=normalfont,labelsep=colon,strut=off} % DO NOT CHANGE THIS
\floatstyle{ruled}
\newfloat{listing}{tb}{lst}{}
\floatname{listing}{Listing}
\title{Sequence Generation with Label Augmentation for Relation Extraction}

\author{
Bo Li$^{1,2}$ \equalcontrib, 
Dingyao Yu$^{1,2}$ \equalcontrib, 
Wei Ye$^{1}$\thanks{{} {} Corresponding author.}, 
Jinglei Zhang$^{1,2}$, 
Shikun Zhang$^{1\dagger}$ \\}

\affiliations{
$^1$National Engineering Research Center 
  for Software Engineering, Peking University \\
  $^2$School of Software and Microelectronics, Peking University\\
  {deepblue.lb@stu.pku.edu.cn}, {yudingyao@pku.edu.cn}, {wye@pku.edu.cn},\\ {jinglei.zhang@stu.pku.edu.cn}, {zhangsk@pku.edu.cn}
}

\usepackage{bibentry}
\begin{document}

\maketitle

\begin{abstract}

Sequence generation demonstrates promising performance in recent information extraction efforts, by incorporating large-scale pre-trained Seq2Seq models. This paper investigates the merits of employing sequence generation in relation extraction, finding that with relation names or synonyms as generation targets, their textual semantics and the correlation (in terms of word sequence pattern) among them affect model performance. We then propose \textbf{R}elation \textbf{E}xtraction with \textbf{L}abel \textbf{A}ugmentation (\texttt{RELA}), a Seq2Seq model with automatic label augmentation for RE. By saying label augmentation, we mean prod semantically synonyms for each relation name as the generation target. Besides, we present an in-depth analysis of the Seq2Seq model's behavior when dealing with RE. Experimental results show that \texttt{RELA} achieves competitive results compared with previous methods on four RE datasets \footnote{Code is available at  https://github.com/pkuserc/RELA}.

\end{abstract}

\section{Introduction}\label{intro}
Paradigm shift has been observed in an increasing number of tasks in recent years, that is some paradigms have the potential to solve diverse NLP tasks \cite{DBLP:journals/corr/abs-2109-12575}. For example, although most natural language understanding (NLU) tasks treat the classification-based methods as the default solution, sequence-to-sequence (Seq2Seq) models could achieve on-pair performances \cite{DBLP:conf/acl/YanGDGZQ20,DBLP:conf/acl/0001LDXLHSW22,DBLP:conf/acl/SaxenaKG22,DBLP:conf/acl/Zhang0TW022,DBLP:journals/ijautcomp/SunLQH22,DBLP:conf/acl/MaoSYZC22}. With the help of pre-trained Seq2Seq models \cite{lewis-etal-2020-bart,DBLP:journals/jmlr/RaffelSRLNMZLL20} and carefully designed schema, these methods directly generate various objectives, such as sentiment polarity, class name, entity type, etc. 

There are also numerous works incorporating Seq2Seq models for relation extraction (RE). Previous efforts mainly focus on the following three aspects. The ﬁrst one designs novel modules or model architectures, for example, \citet{DBLP:conf/acl/0002Z022} proposes a framework of text and graph to learn relational reasoning patterns for relational triple extraction. The second one utilizes various external datasets or resources to enrich the input \cite{DBLP:conf/emnlp/CabotN21,DBLP:journals/corr/abs-2206-05123,DBLP:conf/iclr/PaoliniAKMAASXS21,DBLP:conf/acl/0001LDXLHSW22}.  The last one aims to design proper generation schema. For example, \citet{DBLP:conf/naacl/JosifoskiCPP022} leveraged constrained beam search \cite{DBLP:conf/nips/SutskeverVL14,DBLP:conf/iclr/CaoI0P21} to generate reasonable outputs. \citet{DBLP:conf/acl/0001LDXLHSW22} proposed a structured extraction language to better guide the Seq2Seq model. By improving the model itself or enriching the input, the above methods could better utilize the knowledge existing in Seq2Seq model. A natural question is raised: does the target side could also active the knowledge and have significant impact on the Seq2seq-based RE model? In this research, we work towards closing the above research gap by enriching the information existing in relation names.

\begin{figure}[t]
	\centering
	\includegraphics[width=0.8\linewidth]{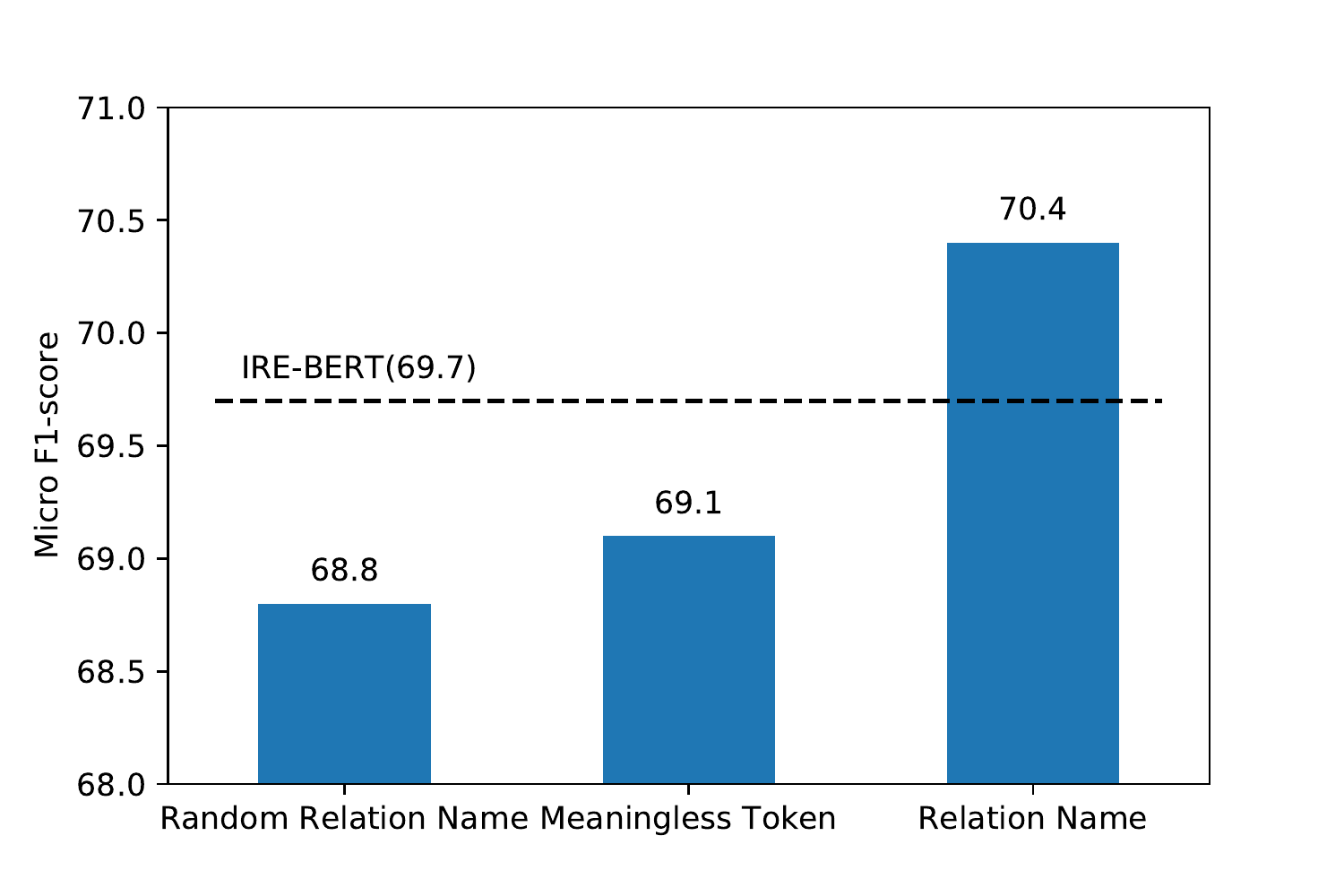}
	\caption{The results of different models on the TACRED test set. We report three kinds of BART-based variants with different generation objectives. The horizontal line is the result from IRE-BERT\cite{DBLP:journals/corr/abs-2102-01373}, which is a strong classification-based model. Note that we use BART-large and BERT-large as backbone networks. The results indicate that the generation target significantly affects the performance of Seq2Seq-based RE model.}
	\label{pilot}
\end{figure}

To answer the aforementioned questions, we first conduct a pilot experiment to explore whether Seq2Seq models could learn useful information from label spaces. Specifically, for a given input and two target entities, we force the BART to generate three kinds of labels: 1) \textbf{Relation Name}. We use original relation names as generation objectives directly. These labels \textit{contain correct information}. 2) \textbf{Meaningless Token}. As its name implies, we replace relation names with some meaningless token, such as ö. These tokens \textit{do not include any useful information}. 3) \textbf{Random Relation Name}. We randomly map the original relation name to another relation name one by one and select the mapped one as the generation objective. Thus for a given input, the corresponding relation name \textit{provides misleading information}.\footnote{Details of these label transformations can be found in the Appendix \ref{app.pilot_exp}.} The results are shown in Figure \ref{pilot}, from which we can see that directly generating correct relation names achieves the highest F1-score, while training with misleading label semantic information largely behind. The above results indicate that label information significantly affects Seq2Seq-based RE models, and these correct relation names contain valuable information.

In fact, relation names existing in most public RE datasets are meaningful phrases. These phrases describe corresponding relations and carry more or less available information. Normally, we find that most relation names contain two types of information: 1) \textbf{Semantic Information}. Relation names provide highly abstract semantics since they are mainly composed of nouns and verbs. 2) \textbf{Correlation Information}. Some relations have the same patterns or writing styles, e.g., \textit{place of birth} and \textit{place of death} both have the prefix \textit{place of}. These patterns may exhibit some correlations between relations. We argue that Seq2Seq methods could utilize the above two types of information from relation names and benefit for RE (\S \ref{main_result}).

Based on the above observations, we raise another question: can we enhance the information in relation names, and achieve better performance? With the help of the powerful generative model GPT-2\cite{Radford2019LanguageMA} and off-the-shelf synonym dictionary, in this research, we introduce \textbf{R}elation \textbf{E}xtraction with \textbf{L}abel \textbf{A}ugmentation (\texttt{RELA}). Specifically, we extend the relation names via three approaches: 1) \textbf{Paraphrase} that modifies a given relation name and some prefix phrases as input, then leverages GPT-2 to generate several related words, 2) \textbf{Inquiry} that takes each training text and a query as input and directly outputs the relation name by asking GPT-2, and 3) \textbf{Synonym} that retrievals several synonyms by utilizing an off-the-shelf synonym dictionary. Our experiments show that simply using several synonyms for each relation name could gain remarkable improvements(\S \ref{diff_approach}), especially for the domain-specific RE dataset and the low-resource scenario(\S \ref{low_resource_exp}).

Finally, we investigate the BART's behavior when dealing with classical generation tasks (such as summarization) and relation extraction. Our experiments show that BART exhibits distinctly different behaviors among different tasks. Interestingly, we find that when generating relation names, BART highly relies on the $\langle bos\rangle$ token and almost ignores previous decoder outputs, which is different from classical generation tasks. We report more visualization and in-depth analysis in the following part(\S \ref{analysis}).   
%while solving RE with Seq2Seq does not seem like an actual generation process, this model indeed benefits from the relation names
Below we summarize our main contributions:
\begin{enumerate}[leftmargin=*]
    \item We explore the advantages of utilizing Seq2Seq models for relation extraction task. We find that Seq2Seq models could obtain good performance by leveraging semantic information and correlation information existing in relation names.
    \item We propose \textbf{R}elation \textbf{E}xtraction with \textbf{L}abel \textbf{A}ugmentation (\texttt{RELA}), a simply yet effective method to extend the relation names and achieves desirable performances among four RE datasets.
    \item We present an in-depth analysis of the BART's behavior when dealing with RE and classical generation task and conduct some instructive conclusions, which may be useful for future work.
\end{enumerate}
\section{Approach}
In this section, we first describe the Seq2Seq method for relation extraction. Then, we will introduce three augmentation methods to enrich relation names in detail.

\subsection{Sequence-to-Sequence Relation Extraction Model}

\begin{figure}[t]
	\centering
	\includegraphics[width=0.99\linewidth]{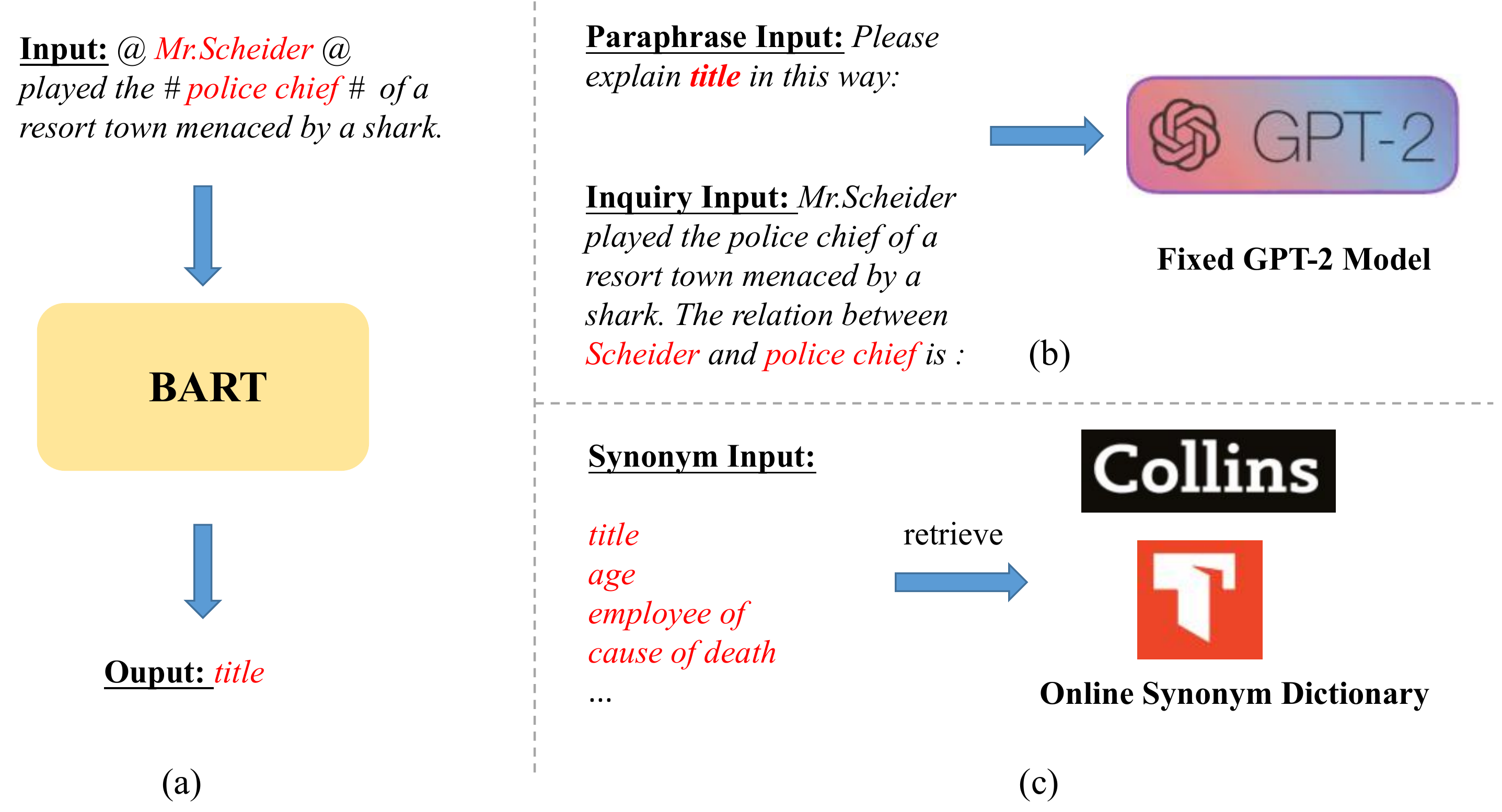}
	\caption{The Sequence-to-Sequence relation extraction model and three label augmentation methods. We take the following instance as an example: \textit{\underline{Mr.Scheider} played the \underline{police chief} of a resort town menaced by a shark}. The target entities are \underline{underlined}, and the ground truth relation is \textit{`title'}. In (a), we fine-tune BART to generate the relation name directly. In (b) and (c), we use GPT-2 and off-the-shelf online synonym dictionaries to augment label information automatically.}
	\label{model}
\end{figure}

For a given input sentence $s$ contains $N$ words, relation extraction aims at extracting the relationship between $h$ and $t$, where $h$ and $t$ are two given target entities in $s$. Standard classification-based RE models transform labels into one-hot vectors with cross-entropy loss \cite{DBLP:conf/acl/MiwaB16,DBLP:conf/acl/GuoZL19,DBLP:conf/acl/YeLXSCZ19,DBLP:conf/acl/SoaresFLK19,DBLP:conf/aaai/LiYHZ21,DBLP:conf/emnlp/RoyP21}. However, these approaches largely ignore both semantic information and correlation information existing in relation names. While relation names are proven to be beneﬁcial for RE in \S \ref{intro}, we introduce the Seq2Seq models to leverage this valuable information via decoding the relation names directly. In fact, some previous works already explored Seq2Seq models for information extraction. They mainly focus on relational triple extraction with large-scale weakly supervised datasets pre-training \cite{DBLP:conf/emnlp/CabotN21,DBLP:conf/iclr/PaoliniAKMAASXS21}. In this research, we argue that it is possible to directly generate relation names for the RE task, as shown in Figure \ref{model} (a).

We treat relation extraction as a generation task following the standard Seq2Seq manner. We first highlight the target entities by enclosing entity spans with two special tokens, e.g., `@' for the head entity and `\#' for the tail entity. The modified input sentence $\hat{s}$ is then passed into the transformer encoder. Finally, we use an auto-regressive transformer decoder to generate the relation name $Y$, where the label length is $L$. Note we generate each token in $Y$ one by one. The training manner is similar to the classical Seq2Seq task, such as the summarization. In this research, we use pre-trained BART as the backbone network. We denote this Seq2Seq-based RE model as \texttt{BART-RE}. By fine-tuning BART on the downstream RE dataset, we minimize the sequential cross-entropy loss between the relation name and the model's output. The training procedure can be written as follows:

\begin{equation}
    p(Y|\hat{s}) = \prod_{t = 1}^{T}p(y_t|\hat{s},y_{<t}),
\end{equation}

\subsection{Relation Extraction with Label Augmentation}\label{rela}
From the pilot experiment in (\S \ref{intro}) we can see that the Seq2Seq-based RE model can learn valuable information from relation names and achieve impressive results. However, relation names usually contain few words, which may hinder BART to fully understand their meanings. Based on this, we argue that enriching the relation names would provide more supplementary information and benefit the model's performance. In this research, we introduce \textbf{R}elation \textbf{E}xtraction with \textbf{L}abel \textbf{A}ugmentation (\texttt{RELA}). By utilizing the powerful generative model GPT-2 \footnote{We use the large version from https://huggingface.co/gpt2.} and off-the-shelf synonym dictionary, as shown in Figure \ref{model}, we propose three automatic label augmentation approaches: \textbf{Paraphrase}, \textbf{Inquiry} and \textbf{Synonym}. 

\begin{table*}[]
\centering
\setlength{\tabcolsep}{2mm}
\begin{tabular}{c|c|c|c}
\hline
\textbf{Relation Name}    & \textbf{Paraphrase} & \textbf{Inquiry}             & \textbf{Synonym}            \\ \hline
\textbf{place of birth}   & born, family        & first appearance, first time & birthplace, born in a place \\ 
\textbf{education degree} & college, student    & law school, from university  & study, graduate             \\ 
\textbf{institution}      & group, community    & high school, new york        & institution, organization,  \\ 
\textbf{date of birth}    & birth date, born     & years old, one founders      & birthday, time of birth,    \\ 
\textbf{place of death}   & death place, kill   & died on, who died            & dead in a palce, deathplace, \\ \hline
\end{tabular}
\caption{Augmented information for each relation name via different approaches.}
\label{googlere_example}
\end{table*}

\begin{enumerate}[leftmargin=*]
    \item \textbf{Paraphrase.} For a given relation name $r$, \textbf{Paraphrase} first converts $r$ as a query by adding a prefix and then feed to the GPT-2 to explain the meaning of $r$. For example, relation name \textit{`title'} will be transformed to \textit{`Please explain title in this way: '}. To reduce the randomness with considering the variety, we use five different prefixes and run ten times with different random seeds for each of them. The above process can generate diverse and complementary explanations for the given relation name. After explaining all relation names, we use TF-IDF to distinguish the top-50 most related phrases for each relation. Finally, we manually select two most relevant phrases for each relation name as the augmented information.
    
    \item \textbf{Inquiry.} This method directly uses the input text and a question to guide the GPT-2 output the target relation. For example, for the given input, \textit{`Mr.Scheider played the police chief of a resort town menaced by a shark'}. The two target entities are\textit{'Scheider'} and \textit{'police chief'}. We use the following question to ask the GPT-2: \textit{`Mr.Scheider played the police chief of a resort town menaced by a shark. The relation between Scheider and police chief is :'}.\footnote{The ground truth is 'title'} We run fifty times for each input with different random seeds to generate more answers. Note that we only collect answers from the training set. We gather all answers for each relation name and form a long text, then use TF-IDF and the same selection process in \textbf{Paraphrase} to choose two most relevant phrases for each relation name.
    
    \item \textbf{Synonym.} Although the above two approaches can generate some valuable information for relation names, useless information is unavoidable in the meanwhile. To further improve the quality of the augmented information, we utilize the off-the-shelf online synonym dictionary, e.g., collins dictionary\footnote{https://www.collinsdictionary.com/} and thesaurus. \footnote{https://www.thesaurus.com/} We directly retrieve several synonyms for each relation name as the augmented information.
    
\end{enumerate}

The final generation objective is the concatenation of the original relation name and its corresponding augmented information. For example, given a relation name \textit{`place of birth'} and its two synonyms \textit{`birthplace'} and \textit{`born in a place'}, the generation target for \texttt{RELA} is \textit{`place of birth, birthplace, born in a place'}. When decoding, only exactly generating \textit{`place of birth, birthplace, born in a place'} (instead of \textit{`place of birth'}) can yield a correct relation prediction. Any mismatch (including OOV) will be treated as \textit{`no relation'}. \texttt{RELA} can be viewed as an extension of the \texttt{BART-RE} model. Due to the space limitation, we only show the detailed augmented information for the Google RE dataset in Table \ref{googlere_example}. Others are shown in the Appendix \ref{app.label_transformation}.

\section{Experiments}

\subsection{Datasets}

We evaluate our method on four RE datasets, i.e., TACRED \cite{DBLP:conf/emnlp/ZhangZCAM17}, SemEval2010 \cite{DBLP:conf/semeval/HendrickxKKNSPP10}, Google RE,\footnote{https://github.com/google-research-datasets/relation-extraction-corpus} and sciERC \cite{DBLP:conf/emnlp/LuanHOH18}. Specifically, TACRED, SemEval2010, and Google RE are commonly used in previous works. To verify the generalization ability in the domain-specific scenario, we also consider using sciERC in our experiments, a large RE dataset with scientific entities and their relations. Table \ref{data} shows the detailed datasets information used in this research. Note that we use the modified version released by \citet{DBLP:conf/emnlp/BeltagyLC19}, because it removes lots of redundant sentences. We use the Micro-F1 score among positive relation labels (excluding `no relation') as the evaluation metric. Following previous works \cite{DBLP:conf/emnlp/YamadaASTM20,DBLP:conf/emnlp/CabotN21}, we do not include any entity type information, because the annotation of entity type is frequently violated in practice.
 
\begin{table}[h]
\centering
\setlength{\tabcolsep}{3mm}
\begin{tabular}{l|c|c|c|c}
\hline
Dataset   & \#Train & \#Dev  & \#Test & \#Class \\ \hline
TACRED    & 68,124  & 22,613 & 15,509 & 42      \\
SemEval   & 8,000   & -      & 2,712  & 10      \\
Google RE & 38,112   &  9,648  &  9,616  & 5      \\
sciERC    &  3,219   &  455   &  974   & 7       \\ \hline
\end{tabular}
\caption{Statistics of different RE datasets used in our experiments. SemEval2010 does not have validation set. \#classes contains the `No Relation' instances.}
\label{data}
\end{table}

\subsection{Experimental Setup}
We use Pytorch \cite{DBLP:conf/nips/PaszkeGMLBCKLGA19} and Tesla T4 GPU with a batch size of 8. For \texttt{RELA} and its variants, we use BART-large as the backbone network, and the checkpoint is downloaded here.\footnote{https://huggingface.co/facebook/bart-large} The maximum input/output sequence length is 256/32 tokens. As an optimiser, we used AdamW \cite{DBLP:conf/iclr/LoshchilovH19} with a 1e-5 learning rate and a 0.2 warmup ratio. The training epoch is 20 for Semeval and 10 for other datasets. We choose the checkpoint that achieves the best result on the development set and evaluates the test set. To reduce the randomness, we run each model five times in the standard scenario and ten times in the low-resource scenario.

\begin{table*}[h]
\centering
\setlength{\tabcolsep}{4mm}
\begin{tabular}{c|c|c|c|c|c|c}
\toprule[1.5pt]
\textbf{Method}       & \textbf{\begin{tabular}[c]{@{}c@{}}Semantic\\ Information\end{tabular}} & \textbf{\begin{tabular}[c]{@{}c@{}}Correlation\\ Information\end{tabular}} & \textbf{TACRED}     & \textbf{SemEval}    & \textbf{Google RE}  & \textbf{sciERC}     \\ \hline
\textbf{IRE-BERT}    & \XSolidBrush                                                                 & \XSolidBrush                                                                & 69.7                & 89.1                & 92.2                & 88.8                \\ 
\textbf{IRE-RoBERTa} & \XSolidBrush                                                                 & \XSolidBrush                                                                & 70.5                & 89.8                & 93.1                & 88.9                \\ 
\textbf{MTB} & \XSolidBrush                                                                 & \XSolidBrush                                                                & 71.5                & 89.5                & 92.7                & 87.4                \\ 
\textbf{LUKE} & \XSolidBrush                                                                 & \XSolidBrush                                                                & \underline{72.7}                & 90.1                & \underline{94.0}                & 87.7                \\\hline 
\textbf{REBEL}      & \checkmark                                                                 & \checkmark                                                                &  70.7    &   82.0     &   93.5        &   86.3    \\ 
\textbf{TANL(T5)}      & \checkmark                                                                 & \checkmark                                                                &  71.9    &   -     &   -        &   -    \\ \hline
\texttt{BART-RE}      & \checkmark                                                                 & \checkmark                                                                & 70.4(-0.1)          & 89.7(-0.1)          & 93.3(+0.2)          & 88.8(-0.1)          \\ 
\texttt{BART-DS}      & \XSolidBrush                                                                 & \checkmark                                                                & 69.3(-1.2)          & 89.4(-0.4)          & 93.1(+0.0)          & 88.3(-0.6)          \\ 
\texttt{BART-DC}      & \checkmark                                                                 & \XSolidBrush                                                                & 70.1(-0.4)          & -    & 93.0(-0.1)          & 88.6(-0.3)          \\ 
\texttt{BART-DB}      & \XSolidBrush                                                                 & \XSolidBrush                                                                & 69.1(-1.4)          & 89.4(-0.4)          & 92.2(-0.9)          & 88.1(-0.8)          \\ 
\texttt{RELA} & \checkmark                                                                 & \checkmark                                                                & \textbf{71.2(+0.7)} & \textbf{90.4(+0.6)} & \textbf{93.9(+0.8)} & \textbf{90.3(+1.4)} \\ \bottomrule[1.5pt]
\end{tabular}
\caption{Performance of different methods on four relation extraction datasets. Results are all cited from public papers or re-implemented with official open-source code. \texttt{RELA} here is the implementation of \texttt{BART-Synonym}. \underline{Underlined} results are previous state-of-the-art, \textbf{bold} results are the best performances \texttt{RELA} achieved. We left blank in \texttt{BART-DC} on the SemEval dataset since relation names in this dataset do not contain any similar writing style. Results of our methods are averaged over five random seeds, and the results are statistically significant with $p < 0.05$. We use the large version of BERT, RoBERTa, and BART in our experiments.}
\label{main}
\end{table*}

\subsection{Comparison Models}\label{comparison_model}

Three types of models are considered in our comparison: 1) Classification-based models, LUKE \cite{DBLP:conf/emnlp/YamadaASTM20}, MTB \cite{DBLP:conf/acl/SoaresFLK19} and IRE \cite{DBLP:journals/corr/abs-2102-01373}, 2) Seq2Seq-based models, REBEL \cite{DBLP:conf/emnlp/CabotN21} and TANL \cite{DBLP:conf/iclr/PaoliniAKMAASXS21}, and 3) several variants of our models. 

\begin{enumerate}[leftmargin=*]
   \item \textbf{Classification-based Model.} LUKE is a large pre-trained language model with external entity-annotated corpus, which achieves impressive results on several information extraction tasks. MTB leveraged entity linking datasets to generate better entity and relation representations. IRE is a powerful RE model without external corpus usage or additional pre-training. %SciBERT is a domain-specific pre-trained model designed for scientific information extraction tasks.
    \item \textbf{Seq2Seq-based Model.} REBEL pre-trained a BART-large model with a huge external corpus for relational triplet extraction. We download the pre-trained model from the open-source code,\footnote{https://github.com/Babelscape/rebel} and then fine-tune REBEL to directly generate relation names with the same input format as our method. As for TANL, it utilized T5 as the backbone network and took structured prediction as a translation task.
    \item \textbf{Variants of Our Model.} We first build a BART model that directly generates relation names, named as \texttt{BART-RE}. To explore which information BART learns from relation names, we then design three kinds of variants: 1) \texttt{BART-DS}, which \textbf{D}rops the \textbf{S}emantic information in relation names while keeping their label relevance. Specially, every token in the label space will be transformed into a special meaningless token, which is a rare token from BART vocabulary. For example, \textit{place of birth} will be transformed as \textit{)+ cffffcc û}, and \textit{place of death} will be transformed as \textit{)+ cffffcc {]}{[}}. 2) \texttt{BART-DC} is designed to keep semantic information while \textbf{D}rop the \textbf{C}orrelation information between relation names. We achieve this by replacing each relation name with its synonym while eliminating the correlation information. 3) \texttt{BART-DB}, this variant \textbf{D}rops \textbf{B}oth the semantic information and the label relevance within relation names. We achieve this by mapping each relation name into a unique meaningless token. In addition, to justify the effectiveness of the enhancement in relation names, we also test the performances of three different methods introduced in (\S \ref{rela}), which are \texttt{BART-Paraphrase}, \texttt{BART-Inquiry}, and \texttt{BART-Synonym}. The detail of these training objectives can be found in the Appendix \ref{app.label_transformation}. We choose two augmented phrases for each relation name, and the ablation study on the effectiveness of selection numbers can be found in the Appendix \ref{app.number_exp}. 
\end{enumerate}

% \usepackage{multirow}
% \usepackage{multirow}

\subsection{Main Results}\label{main_result}
Table \ref{main} shows the main results on the test sets. The first block exhibits classification-based models, the second block shows Seq2Seq-based models, and the last block is our method with several model variants. Rather than propose a new state-of-the-art method, we intend to propose a simple Seq2Seq model for RE in this research. We mainly compare our approach with strong RoBERTa-based models without using any external pre-training corpus, thus numbers in parentheses are improvements compared with IRE-RoBERTa. From Table \ref{main} we can see that:
\begin{enumerate}[leftmargin=*]
    \item We find that \texttt{BART-RE} is a strong model equipped with Seq2Seq architecture. It performs on par or even better with IRE-RoBERTa. The above results indicate that directly generating relation names is an reliable solution for RE. Compared with three types of variants, ignoring semantic information (\texttt{BART-DS}) or correlation information (\texttt{BART-DC}) decreases the model's performance. Semantic information is more crucial than correlation information in terms of performance degradation. \texttt{BART-DB} has the worst performance since it drops the above two types of information. These results show that the Seq2Seq method learns both semantic information and correlation information existing in relation names, and achieves competitive performances compared with previous strong baseline models.
    \item After enriching relation names, \texttt{RELA} obtains consistent improvements and achieve much better results than IRE-RoBERTa. We attribute this to \texttt{RELA} could learn more useful information from enhanced relation names. Interestingly, for the domain-specific dataset sciERC, \texttt{RELA} gains 1.4\% F1-score improvements over IRE-RoBERTa, which is much bigger than the rest of the datasets. The lesson here is that \texttt{RELA} is more robust via label augmentation when original relation names are difficult to understand. This conclusion also can be learned from the comparison with LUKE. Although LUKE is pre-trained with a large external entity-annotated corpus, which helps LUKE to be more powerful on the dataset with commonly used relations, \texttt{RELA} outperforms LUKE by a large margin when dealing with domain-specific datasets, e.g., +2.9\% F1-score on sciERC. The above results demonstrate that although the original relation name contains valuable information, it may not be enough for Seq2Seq methods. With the help of automatic label augmentation, \texttt{RELA} could strengthen the expressive power of Seq2Seq methods. 
    \item Compared with previous Seq2Seq-based RE architectures, although REBEL was pre-trained with a large external relational triplet extraction dataset, \texttt{RELA} still surpasses REBEL among all datasets. We guess pre-trained REBEL may not be suitable to generate relation names directly. TANL(T5) achieves better results on TACRED than \texttt{RELA}. However, it needs manually designed prompt and utilizes T5 as the backbone network, and the model size is much larger than RoBERTa-large or BART-large. To sum up, \texttt{RELA} achieves comparable results compared with previous excellent methods and can be trained with an academic budget. While being slightly worse than some techniques that are pre-trained with large external datasets or larger models, \texttt{RELA}'s simplicity and effectiveness make it an ideal Seq2Seq baseline for relation extraction.      
\end{enumerate}

%Comparing with LUKE, RELA doesn't achieve better results on TACRED and Google RE. We think pre-training with a large external entity-annotated corpus helps LUKE more powerful on the dataset with common relation names. Even so, RELA outperforms all other models when dealing with relation names that are hard to understand (e.g., SemEval and sciERC).

\begin{figure}[h]
    \centering
    \includegraphics[width=0.45\textwidth]{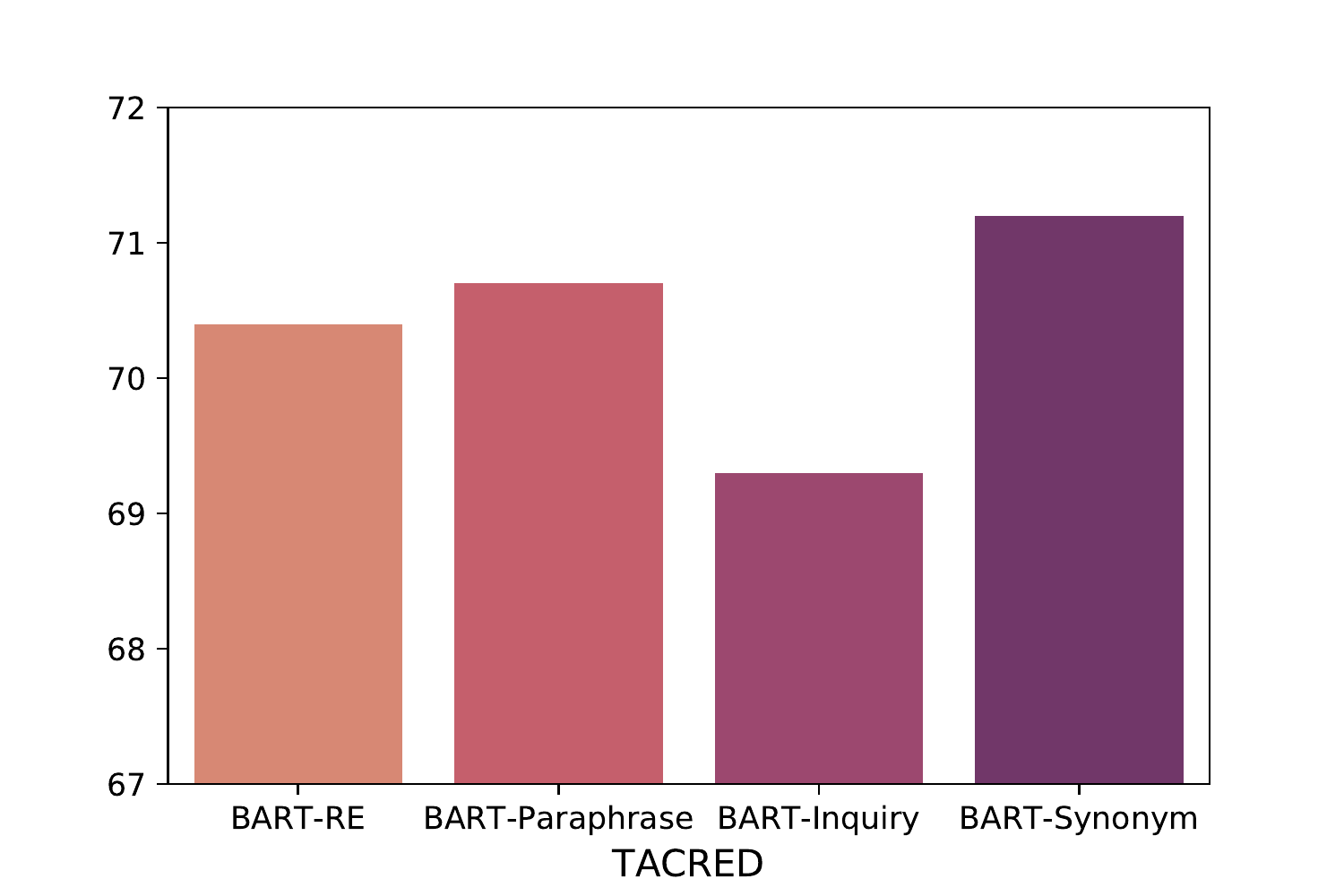}
    \caption{The Effectiveness of different Relation Name Extension Approaches on TACRED. Due to the space limitation, we only report the results on TACRED here. Full results on four datasets can be found in the Appendix \ref{app.approach}}
    \label{approach}
\end{figure}

\subsection{The Effectiveness of different Relation Name Extension Approaches}\label{diff_approach}

We explore the performances of three different automatic label augmentation approaches in this subsection. Figure \ref{approach} shows that: 1) \textbf{Synonym} consistently outperforms other approaches and achieves remarkable improvements over \texttt{BART-RE}. 2) The improvements obtained from \textbf{Paraphrase} or \textbf{Inquiry} are mixed. \textbf{Paraphrase} could provide useful label augmentation information to some extent, while \textbf{Inquiry} is not. With detailed augmented information from all approaches,\footnote{Reader can find details in Appendix \ref{app.approach}} we find that \textbf{Synonym} could provide semantically close and correct supplementary information for each relation name. At the same time, \textbf{Paraphrase} and \textbf{Inquiry} usually generate some irrelevant information, which may not always be helpful for the Seq2Seq-based RE model. Insightfully, we notice that \textbf{Paraphrase} and \textbf{Inquiry} sometimes imitate associative thinking of human, generating some supplementary information like \textit{`grew up'} from original relation name \textit{`city of birth'} and \textit{`real name'} from original relation name \textit{`person alternate name'}. Unfortunately, this supplementary information may not always useful in the current scenario. To sum up, all augmentation methods could provide some supplementary information, and the \textbf{Synonym} is the most stable and effective one.

\begin{table}[h]
\centering
\setlength{\tabcolsep}{1.5mm}
\begin{tabular}{c|c|c|c}
\hline
\textbf{\textit{n}} & IRE-RoBERTa & \texttt{BART-RE} & \texttt{RELA} \\ \hline
\textbf{8}      &  60.3  &  66.7   &  72.3 (+12.0)      \\ \hline
\textbf{16}     &  69.8  &  71.1   &  78.3 (+8.5)         \\ \hline
\textbf{32}     &  73.4  &  73.8   &  80.5 (+7.1)     \\ \hline
\textbf{64}     &  82.1  &  81.7   &  85.6 (+3.5)       \\ \hline
\end{tabular}
\caption{Low-resource Setting. Due to the space limitation, we only report results on the Google RE dataset, and similar trends emerge in other datasets. \textit{n} means the sample number for each label. We randomly select training samples ten times and report the mean F1-score on the whole test set to reduce the randomness.}
\label{low_resource}
\end{table}

\subsection{Low-resource Setting}\label{low_resource_exp}
Since \texttt{RELA} could learn more helpful information from automatic label augmentation, we believe \texttt{RELA} is more suitable for the low-resource setting. To verify our assumption, we conduct low-resource experiments on Google RE. We vary the number of training samples ranging from 8 to 64 for each label, and the comparison results with IRE-RoBERTa are shown in Table \ref{low_resource}. We can observe that \texttt{RELA} outperforms IRE-RoBERTa by a large margin in all settings. These results indicate that our method has significant advantages for the low-resource setting. Besides, \texttt{RELA} is especially effective when the training sample size is very tiny, i.e., when each class only has 8 samples, \texttt{RELA} achieves 12.0\% F1-score improvements compared with IRE-RoBERTa. We believe the label augmentation methods provide supplementary information for \texttt{RELA} when dealing with low-resource RE.

\section{Analysis}\label{analysis}

In this research, we train our model in a Seq2Seq manner by generating relation names one by one. We are interested in the BART's behavior when dealing with different scenarios, i.e., relation extraction v.s. summarization. We visualize several training details on BART's hidden states and conduct some instructive conclusions. We find that: 1) BART exhibits totally different behaviors when dealing with RE and classical sequence generation tasks (e.g., summarization). Although trained by sequential cross-entropy loss, \texttt{BART-RE} and \texttt{RELA} are more similar to the classification method. 2) \texttt{RELA} could learn more separable relation representations than \texttt{BART-RE} via label augmentation.

\subsection{Differences with Classical Sequential Generation Tasks}

\begin{figure}[ht]
    \centering
    \includegraphics[width=0.4\textwidth]{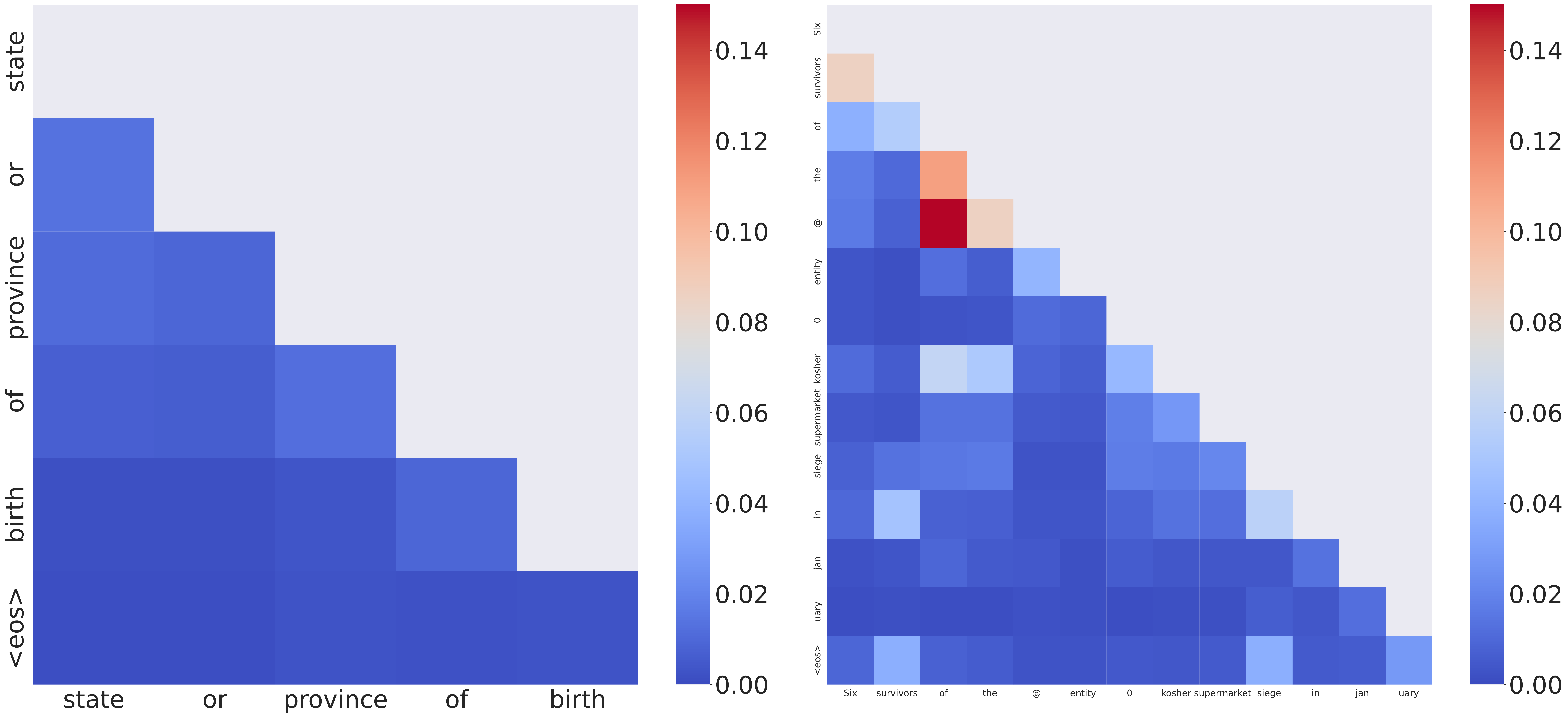}
    \caption{Average decoder self-attention weights among attention heads of different decoding steps (except $\langle bos\rangle$ tag). Left: BART-large for relation extraction; Right: BART-large for text summarization. The self-attention weights of the last decoder layer indicate that \texttt{BART-RE} shows less reliance on previous decoder inputs.}
    \label{fig:img5.1}
\end{figure}

\subsubsection{Dependency on Decoder Inputs}

One of the advantages of sequence generation methods is that previous decoder inputs are visible for the current decoder, thus it is helpful to generate fluent and accurate outputs. But in the case of RE, we find that BART does not pay much attention to previous decoder inputs. Actually, both RE and summarization models spend a high attention score on the $\langle bos\rangle$ tag. In Figure \ref{fig:img5.1} we exclude the attention weight of $\langle bos\rangle$ tag to get a balanced weight distribution. It turns out that the self-attention weights of the \texttt{BART-RE} are much less than those of the summarization model. In another word, for the given relation name \textit{state or province of birth}, when generating the token `birth', \texttt{BART-RE} almost does not rely on previous decoder outputs. Instead, it is mainly dependent on the $\langle bos\rangle$ tag. This result indicates our previous statement that solving RE with Seq2Seq does not seem like an actual generation process. \texttt{BART-RE} just employs the representation of $\langle bos\rangle$ tag as a probe to generate relation names from encoder states.

\begin{figure}[h]
    \centering
    \includegraphics[width=0.4\textwidth]{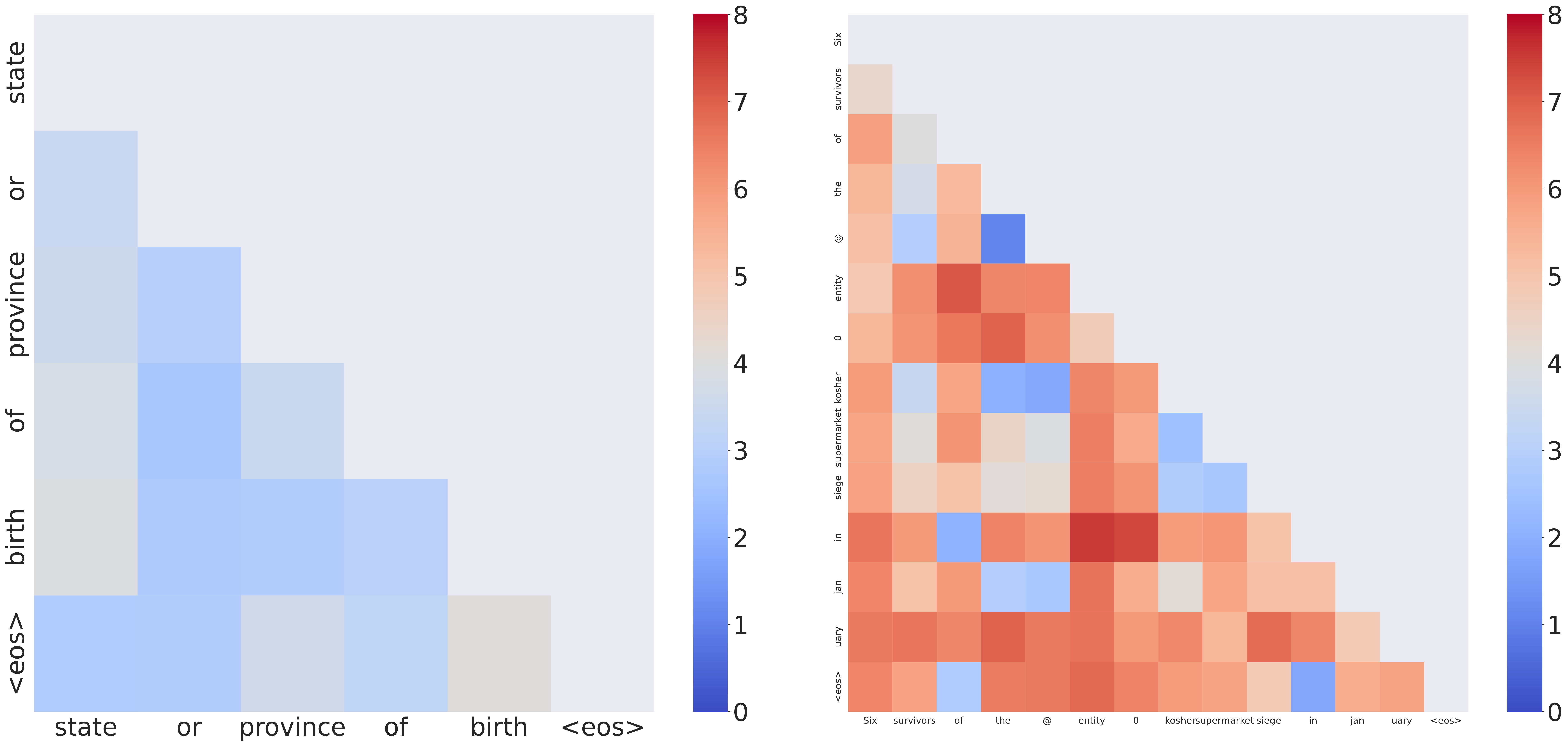}
    \caption{Cross-entropy of the cross-attention weights' distribution among decoding steps. In the \texttt{BART-RE} (Left), low cross-entropy indicates that the distributions of attention weights among different steps contain similar information, while summarization task (Right) has relatively unique cross attention weight distributions.}
    \label{fig:img5.2}
\end{figure}

The above conclusion also can be drawn from another perspective. We explored the cross-attention weights' distribution among each decoding step. In Figure \ref{fig:img5.2}, we employ cross-entropy to measure the similarity of the cross-attention weights' distribution. The results are in line with our expectation. The overall low cross-entropy of \texttt{BART-RE} indicates that the representations obtained from the self-attention mechanism have a high correlation in semantics. These representations do not contain much diverse information for generating relation names.

\begin{figure}[h]
    \centering
    \includegraphics[width=0.4\textwidth]{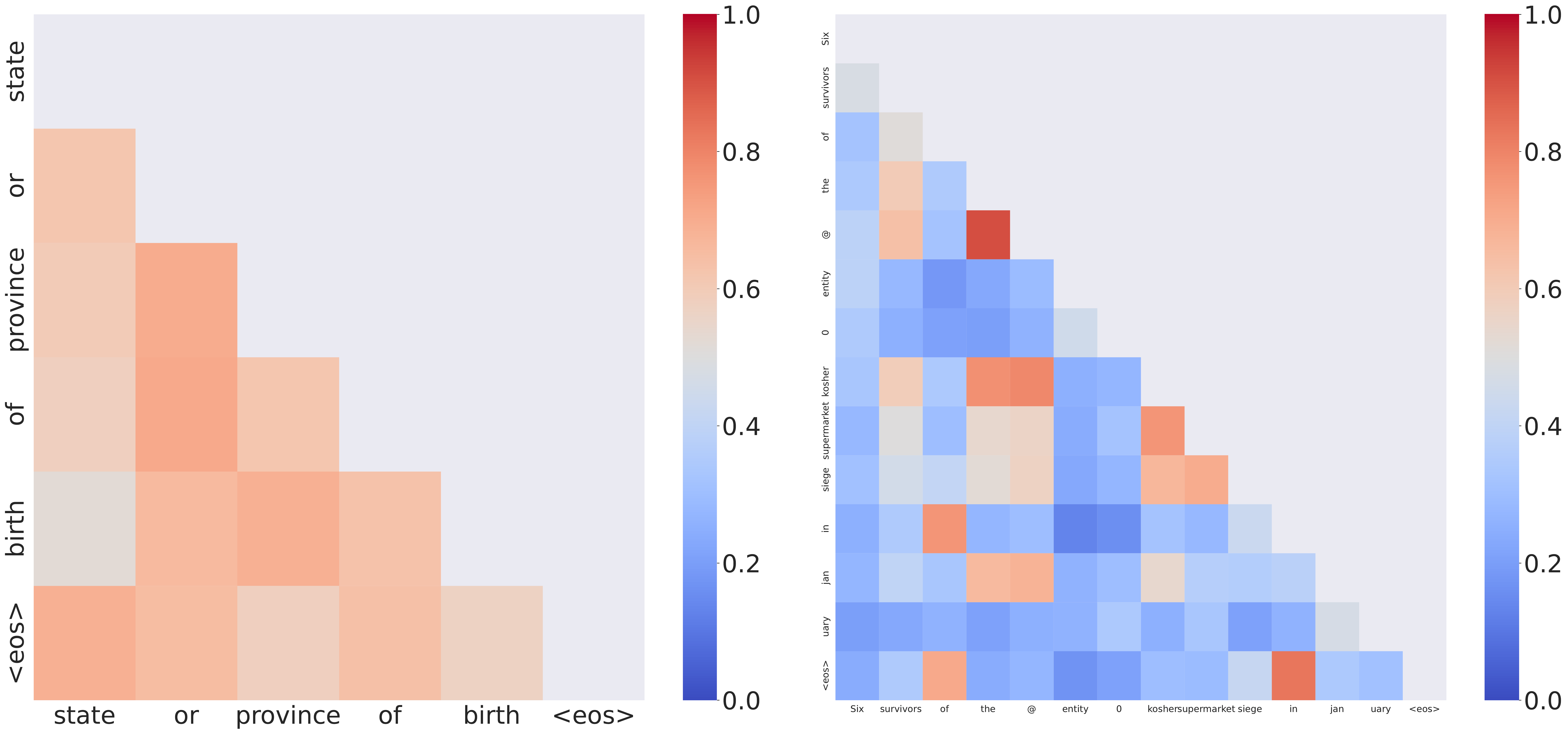}
    \caption{Cosine similarity of the last decoder hidden states between different decoding steps. Left: \texttt{BART-RE}; Right: summarization task. It is complementary to the Figure \ref{fig:img5.2} in the previous subsection. Similar cross-attention weight distributions deduce to similar decoder hidden states.}
    \label{fig:img5.3}
\end{figure}

\subsubsection{Similarity of Decoder Hidden States}

As shown in Figure \ref{fig:img5.3}, corresponding to low cross-entropy among cross-attention weights, the last decoder hidden state of \texttt{BART-RE} has a high score in the case of cosine similarity. Significantly, it is unusual that a preposition "of" and a noun "province" get so close in semantic space. On the other hand, the decoder hidden space is more like a classification space, where the tokens of the same relation belong to the same cluster instead of close in a semantic space.

\subsection{Effects on Decoder Hidden States when Enriching Relation Names}

\begin{figure}[h]
    \centering
    \includegraphics[width=0.4\textwidth]{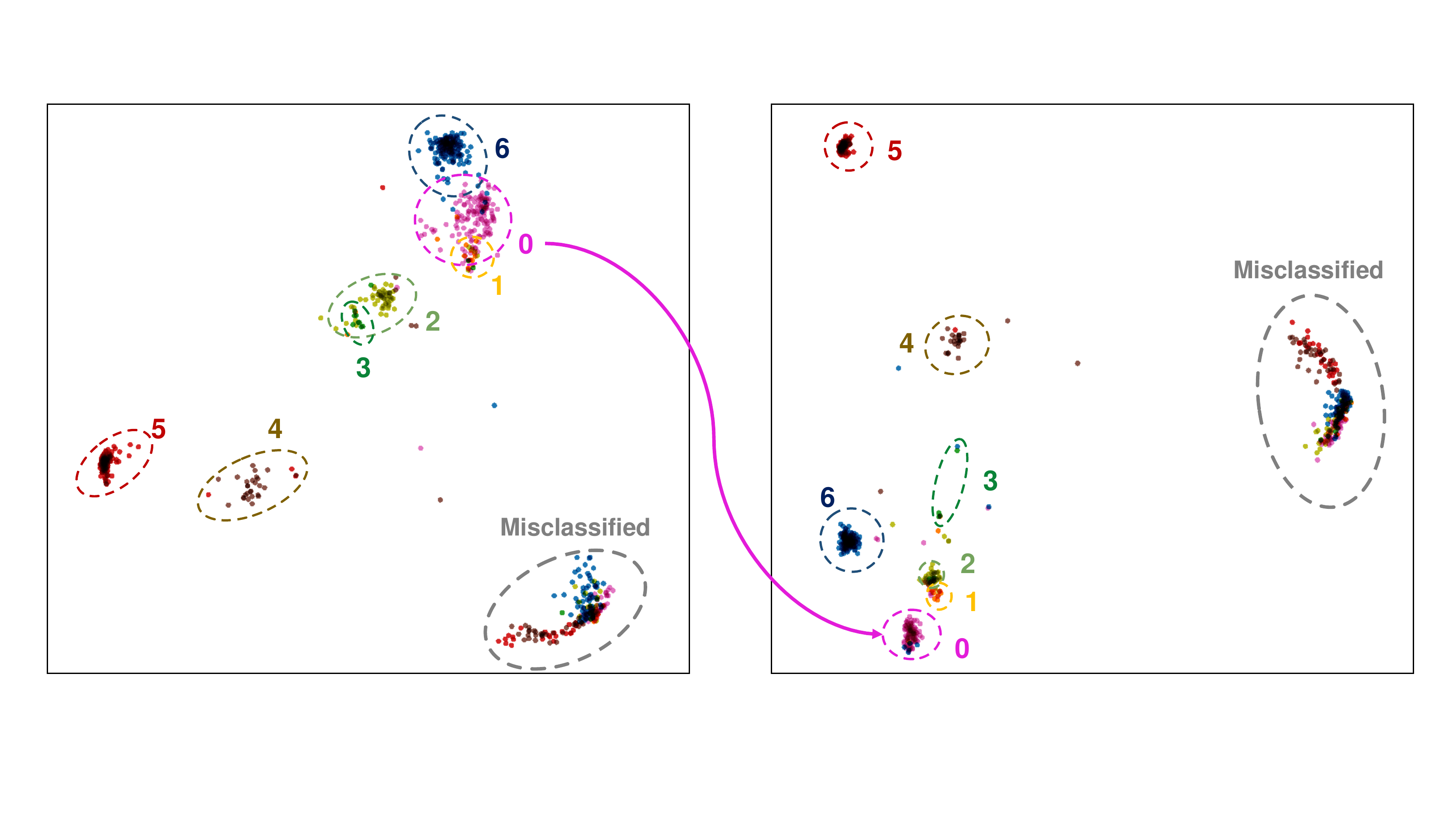}
    \caption{Low dimension representations of average decoder hidden states (Corresponding label names: 0-city of residence, 1-city of death, 2-state or provinces of residence, 3-state or province of birth, 4-organization parent, 5-organization alternate name, 6-employee of). Left: \texttt{BART-RE}; Right: \texttt{RELA}. We find that the representations of different labels are further separated in \texttt{RELA}.}
    \label{fig:img5.4}
\end{figure}

In this subsection, we want to explore the changes in the decoder hidden states after enriching relation names. We visualize the last decoding layer of \texttt{BART-RE} and \texttt{RELA} by applying PCA. From Figure \ref{fig:img5.4}, we can see that \texttt{BART-RE} could distinguish different relation names in the semantic space. Meanwhile, \texttt{RELA} makes different relations more separable. We believe \texttt{RELA} learns more useful information from label augmentation and finally outputs better decoder hidden states.

%The analysis in the previous section ensures that decoder hidden states are qualified indicator to measure the model performance.

\iffalse
\subsection{The role of decoder in the model}

To explore when hidden states become adequate for classification and find out what does decoder do if sequence information is not important while decoding, we investigate the encoder hidden states of texts and decoder hidden states of relations.

The result is shown in Figure \ref{fig:img5.5}. In encoding phrase, average representations of texts perform almost uniform distribution, they still contains a lot of complex semantic information and cannot be used directly for classification. The states migration between decoder layer 1, layer 6 and final layer is a process to make the states gradually linearly separable. The decoding module maps the states from semantic space to classification space.

\begin{figure}[htbp]
    \centering
    \includegraphics[width=0.45\textwidth]{decoder_layers.png}
    \caption{PCA low dimension representations of average hidden states (Different colors refer to different labels). Upper left: The last encoder states; Upper right: The decoder hidden states of decoding layer 1; Down left: The decoder hidden states of decoding layer 6; Down right: The last decoder states. BART decoder extracts vectors suitable for classification from scattered text representations.}
    \label{fig:img5.5}
\end{figure}
\fi
\section{Related Work}
 
Most previous RE models are classification-based approaches. Early works on RE used machine learning with handcraft features or existing NLP tools. These method rely on task-specific features and can not generalize well on differet RE datasets \cite{kambhatla2004combining,boschee2005automatic,qian2008exploiting,Nguyen2009Structures,Nguyen2014Regularization,Zelenko2003Kernel}. Recently, researchers used several deep learning networks for relation extraction, from CNNs and LSTMs \cite{DBLP:conf/acl/MiwaB16,DBLP:conf/acl/SantosXZ15,DBLP:conf/acl/WangCML16,DBLP:conf/acl/GuoZL19,DBLP:conf/acl/YeLXSCZ19} to Transformer-based approaches \cite{DBLP:conf/acl/SoaresFLK19,DBLP:conf/emnlp/YamadaASTM20,DBLP:conf/coling/LiYSXXZ20,DBLP:conf/aaai/XueSZC21,DBLP:journals/corr/abs-2102-01373,DBLP:conf/emnlp/RoyP21}. These methods design novel components based on PLMs or pre-trained language model with external entity-annotated corpus, and chieves impressive results on several relation extraction datasets. In general, These models finally distinguish the relation between target entities via a classification layer. 

Recently, some works have explored Seq2Seq approaches for RE tasks. REBEL \cite{DBLP:conf/emnlp/CabotN21} pre-trained a BART-large model with weakly supervised external datasets, then designed triplets linearization to generate reasonable outputs. REBEL mainly focuses on joint entity and relation extraction. TANL \cite{DBLP:conf/iclr/PaoliniAKMAASXS21} is another great seq2seq-based method based on T5 model. They took structured prediction as a translation task and designed special prompts for different tasks. 

Our research is orthogonal to all the above works. We attempt to solve RE from a sequence-to-sequence perspective by introducing the BART to generate the relation name and its augmented information directly. We want to explore the under-explored problem (Seq2Seq architecture for relation extraction with label augmentation) and provide some insights for future research.

\section{Conclusion} 
In this paper, we first explore the advantages of utilizing sequence-to-sequence models for relation extraction task. We find that Seq2Seq architectures are capable of learning semantic information and correlation information from relation names. Then, we propose \textbf{R}elation \textbf{E}xtraction with \textbf{L}abel \textbf{A}ugmentation (\texttt{RELA}), a simply yet effective method to extend the relation names for better performances. Finally, we present an in-depth analysis of the BART's behavior when dealing with RE and provide insightful conclusions. We hope that all these contents could encourage the community to make further exploration and breakthrough towards better Seq2Seq-based RE models.

\section*{Acknowledgements}
This research is supported by the National Key Research And Development Program of China (No. 2021YFC3340101).

\bibliography{aaai23}
%\nobibliography{aaai23}

\appendix
\clearpage
\section{Appendix}\label{sec:appendix}

\subsection{Label Transformations of TACRED dataset for the Pilot Experiments}\label{app.pilot_exp}
We show different labels for three BART-based models mentioned in \S \ref{intro} in Tabel \ref{pilot_data}. Specifically, \textbf{Relation Name} contains correct information, \textbf{Meaningless Token} does not include any meaningful information, and \textbf{Random Relation name} provides wrong information.

\subsection{Performances with Different Number of Synonyms}\label{app.number_exp}
We also investigate that if adding too more synonyms could achieve better performance. From Table \ref{app.number} we can see that comparing with \textit{l} = 0, adding synonyms could improve model's performance. However, more synonyms are not always better. \textit{l} = 3 is worse than \textit{l} = 2, this probably because too many synonyms may increase decoding difficulty, which can negatively affect the performance. To sum up, adding two synonyms for each relation name is a good choice for all.

\begin{table}[h]
\centering
\setlength{\tabcolsep}{5.5mm}
\begin{tabular}{c|c|c}
\hline
\textbf{\textit{l}}  & \textbf{Google RE}  & \textbf{sciERC}     \\ \hline
\textbf{0} & 93.3                & 88.8                \\ \hline
\textbf{1} & 93.5(+0.2)          & 89.8(+1.0)          \\ \hline
\textbf{2} & \textbf{93.9(+0.6)} & \textbf{90.3(+1.5)} \\ \hline
\textbf{3} & 93.8(+0.5)          & 89.3(+0.5)          \\ \hline
\end{tabular}
\caption{Model performances with different numbers of synonyms. \textit{l} is the number of synonyms per relation name. When \textit{l} = 0, we only use the original relation name as the generation objective, thus this setting is identical to BART-RE.}
\label{app.number}
\end{table}

\begin{figure}[h]
    \centering
    \includegraphics[width=0.45\textwidth]{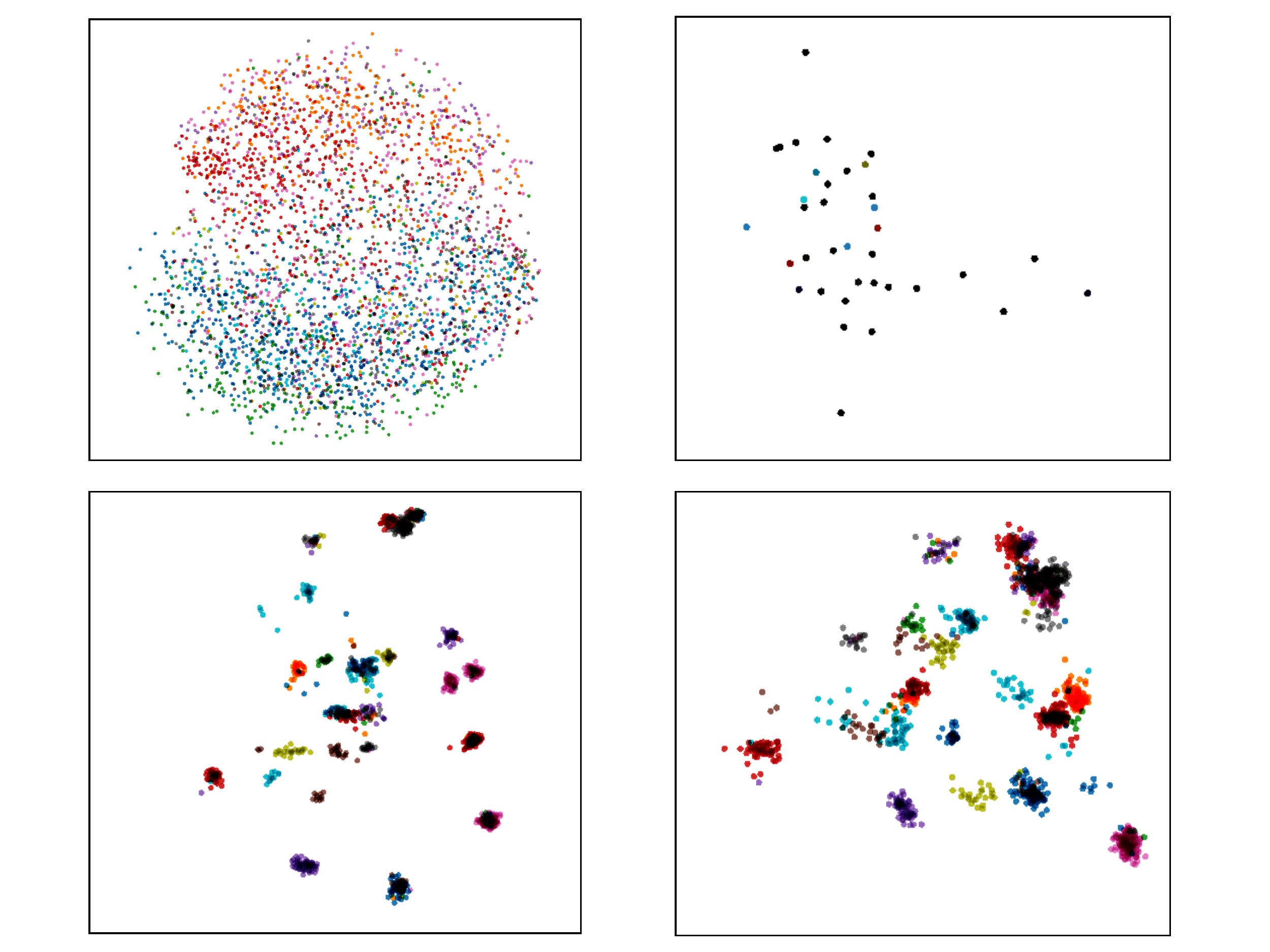}
    \caption{PCA low dimension representations of average hidden states (Different colors refer to different labels). Upper left: The last encoder states; Upper right: The decoder hidden states of decoding layer 1; Down left: The decoder hidden states of decoding layer 6; Down right: The last decoder states. BART decoder extracts vectors suitable for classification from scattered text representations.}
    \label{fig:img5.5}
\end{figure}

\begin{figure*}[t]
    \centering
    \includegraphics[width=0.95\textwidth]{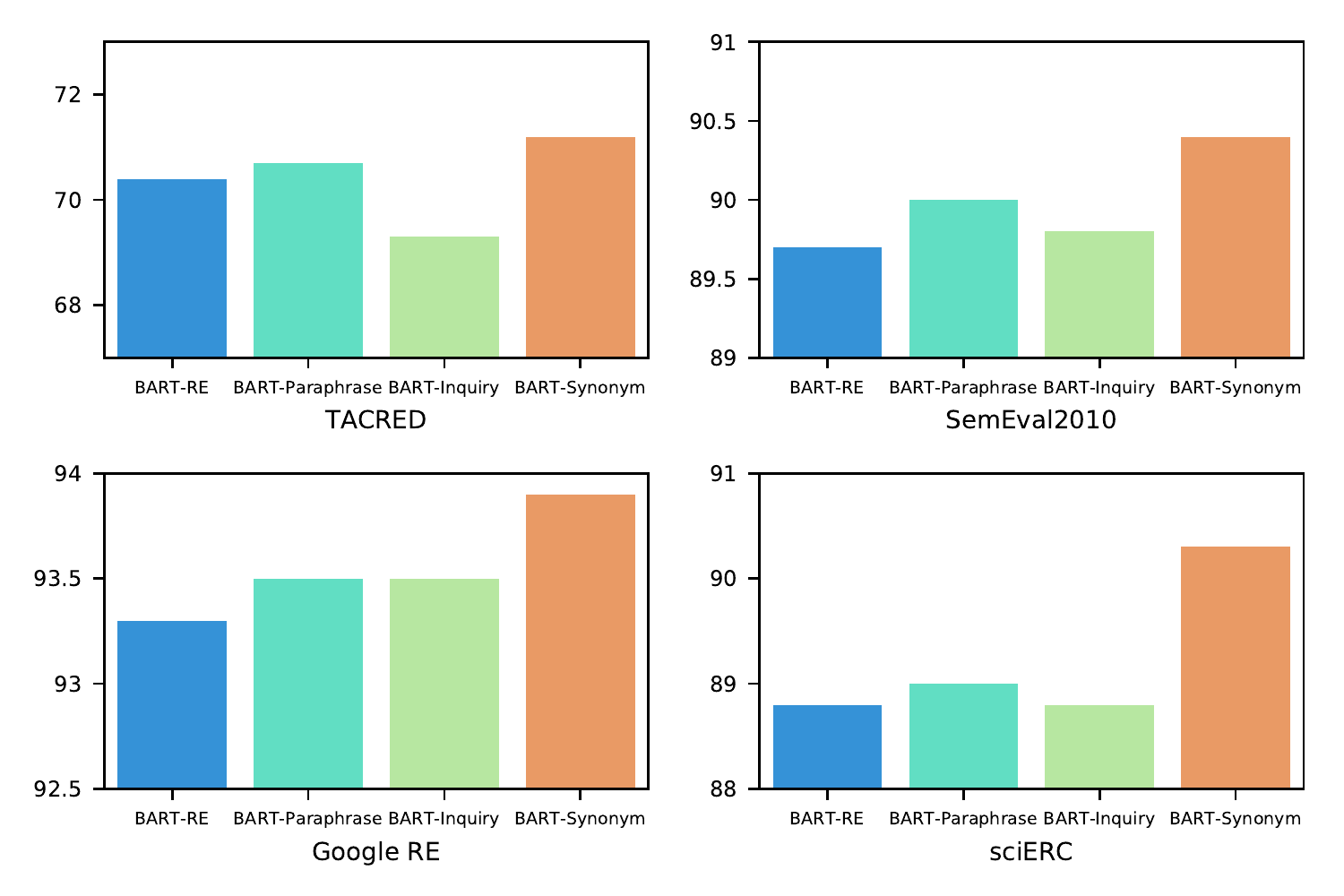}
    \caption{The Effectiveness of different Relation Name Extension Approaches on four datasets. The conclusions here are the same in the \S \ref{diff_approach} }
    \label{app.dff_approach}
\end{figure*}

\subsection{The role of decoder in the model}

To explore when hidden states become adequate for classification and find out what does decoder do while decoding, we investigate the encoder hidden states and decoder hidden states.

The result is shown in Figure \ref{fig:img5.5}. In encoding phase, average representations of texts perform almost uniform distribution, they still contain a lot of complex semantic information and cannot be used directly for classification. The states migration between decoder layer 1, layer 6 and final layer is a process to make the states gradually linearly separable. The decoding module maps the states from semantic space to classification space.

\subsection{The Effectiveness of different Relation Name Extension Approaches}\label{app.approach}

In this subsection, we report the effectiveness of different relation name extension approaches on four RE datasets in Figure \ref{app.dff_approach}. The conclusions here are the same as mentioned in \S \ref{diff_approach}. Synonym performs best among all the approaches, and Paraphrase usually better than Inquiry in most cases. While Inquiry may bring some noise into relation names and may have negative influences on the performance.

\subsection{Different Generation Objectives for Model Variants.}\label{app.label_transformation}
We report the detailed generation objectives for all the model variants on four RE datasets, including BART-RE, BART-DC, BART-DS, BART-DB, BART-Paraphrase, BART-Inquiry and BART-Synonym. We hope these information can inspire researchers to explore more interesting and solid Seq2Seq-based RE systems.

\begin{table*}[ht]
\centering
\begin{tabular}{l|l|l}
\hline
\textbf{Relation Name}   & \textbf{Meanless Token}                                       & \textbf{Random Relation Name}     \\ \hline
date of birth                     & cffff                                                         & religion                          \\
top member employee               & ñ                                                             & cause of death                    \\
no relation                       & \textless{}?                                                  & person parent                     \\
city of residence                 & *=-                                                           & organization alternate name       \\
person parent                     & ö                                                             & country of birth                  \\
country of headquarters           & '\textgreater{}\textgreater{}\textbackslash{}\textbackslash{} & city of headquarters              \\
state or provinces of residence   & ø                                                             & political religious affiliation   \\
other family                      & =$\sim$=$\sim$                                                & state or province of birth        \\
city of birth                     & ú                                                             & charges                           \\
religion                          & û                                                             & city of birth                     \\
members                           & ü                                                             & children                          \\
spouse                            & cffffcc                                                       & dissolved                         \\
number of employees member        & -|                                                            & country of death                  \\
origin                            & \textasciicircum{}\{                                          & title                             \\
title                             & ā                                                             & country of headquarters           \\
political religious affiliation   & ,,,,                                                          & date of birth                     \\
state or province of death        & ||||                                                          & origin                            \\
school attended                   & ():                                                           & state or provinces of residence   \\
cause of death                    & {]}=                                                          & country of residence              \\
country of birth                  & GGGG                                                          & date of death                     \\
siblings                          & ć                                                             & no relation                       \\
employee of                       & ========                                                      & school attended                   \\
city of headquarters              & \_\_\_\_\_\_\_                                                & number of employees member        \\
children                          & 1111                                                          & founded by                        \\
country of death                  & ------                                                        & city of residence                 \\
shareholder                       & (\$                                                           & siblings                          \\
member of                         & ................                                              & website                           \\
organization parent               & ====                                                          & state or province of death        \\
website                           & fff                                                           & subsidiary                        \\
age                               & (-                                                            & state or province of headquarters \\
dissolved                         & .\textgreater{}\textgreater{}                                 & founded                           \\
state or province of birth        & ē                                                             & member of                         \\
subsidiary                        & +++                                                           & other family                      \\
date of death                     & \_-\_                                                         & spouse                            \\
founded by                        & \%{]}                                                         & employee of                       \\
country of residence              & ()\{                                                          & members                           \\
person alternate name             & \$\$\$\$                                                         & city of death                     \\
city of death                     & $\sim \sim \sim \sim$                                      & shareholder                       \\
state or province of headquarters & )+                                                            & age                               \\
organization alternate name       & \textgreater{}:                                               & person alternate name             \\
charges                           & {]}{[}                                                        & organization parent               \\
founded                           & č                                                             & top member employee \\ \hline            
\end{tabular}
\caption{Label transformations of TACRED dataset for the pilot experiments in \S \ref{intro}.}
\label{pilot_data}
\end{table*}

\begin{table*}[]
\centering
\begin{tabular}{l|l|l|l}
\hline
\textbf{BART-RE} & \textbf{BART-DC} & \textbf{BART-DS}       & \textbf{BART-DB}                         \\ \hline
date of birth                         & date of birth                         & (- ā ē                                       & cffff                                                         \\
top member employee                   & top member employee                   & \%{]} ()\{ *=-                               & ñ                                                             \\
no relation                           & no relation                           & ć +++                                        & \textless{}?                                                  \\
city of residence                     & the city where live in                & $\sim\sim\sim\sim$ā cffff              & *=-                                                           \\
person parent                         & the father or mather of               & ==== {]}{[}                                  & ö                                                             \\
country of headquarters               & has its headquarters in               & ö ā \$\$\$\$                                    & '\textgreater{}\textgreater{}\textbackslash{}\textbackslash{} \\
state or provinces of residence       & live in a state or province           & --- fff -| ā cffff                           & ø                                                             \\
other family                          & other family                          & cffffcc \textasciicircum{}\{                 & =$\sim$=$\sim$                                                \\
city of birth                         & was born in a city                    & $\sim\sim\sim\sim$ā ē                  & ú                                                             \\
religion                              & religion                              & \textgreater{}.                              & û                                                             \\
members                               & members                               & ñ                                            & ü                                                             \\
spouse                                & spouse                                & ø                                            & cffffcc                                                       \\
number of employees member            & number of employees member            & 1111 ā |||| ()\{                             & -|                                                            \\
origin                                & origin                                & \_-\_                                        & \textasciicircum{}\{                                          \\
title                                 & title                                 & \_\_\_\_\_\_\_                               & ā                                                             \\
political religious affiliation       & political religious affiliation       & (\$ č .\textgreater{}\textgreater{}          & ,,,,                                                          \\
state or province of death            & died in a state or town               & --- fff {]}= ā ü                             & ||||                                                          \\
school attended                       & school attended                       & ,,,, )+                                      & ():                                                           \\
cause of death                        & cause of death                        & ): ā ü                                       & {]}=                                                          \\
country of birth                      & nationality                           & ö ā ē                                        & GGGG                                                          \\
siblings                              & siblings                              & ------                                       & ć                                                             \\
employee of                           & employee of                           & *=- ā                                        & ========                                                      \\
city of headquarters                  & located in                            & $\sim\sim\sim\sim$ā \$\$\$\$              & \_\_\_\_\_\_\_                                                \\
children                              & children                              & \textgreater{},                              & 1111                                                          \\
country of death                      & died in a country                     & ö ā ü                                        & ------                                                        \\
shareholder                           & shareholder                           & \textgreater{}\textgreater{}\textbackslash{} & (\$                                                           \\
member of                             & member of                             & ()\{ ā                                       & ................                                              \\
organization parent                   & parent company                        & (): {]}{[}                                   & ====                                                          \\
website                               & website                               & =$\sim$=$\sim$                               & fff                                                           \\
age                                   & age                                   & \textgreater{}:                              & (-                                                            \\
dissolved                             & dissolved                             & ú                                            & .\textgreater{}\textgreater{}                                 \\
state or province of birth            & the birth place is a state or town    & --- fff {]}= ā ē                             & ē                                                             \\
subsidiary                            & subsidiary                            & GGGG                                         & +++                                                           \\
date of death                         & the time of when dead                 & (- ā ü                                       & \_-\_                                                         \\
founded by                            & founded by                            & û \textless{}?                               & \%{]}                                                         \\
country of residence                  & residence country                     & ö ā cffff                                    & ()\{                                                          \\
person alternate name                 & person alternate name                 & ==== ................ --                     & \$\$\$\$                                                         \\
city of death                         & city of death                         & $\sim\sim\sim\sim$ā ü                  & $\sim\sim$$\sim$$\sim$                                      \\
state or province of headquarters     & state or province of headquarters     & --- fff {]}= ā \$\$\$\$                         & )+                                                            \\
organization alternate name           & is also known as                      & (): ................ --                      & \textgreater{}:                                               \\
charges                               & charges                               & ========                                     & {]}{[}                                                        \\
founded                               & was formed in                         & û                                   & č  \\\hline                                                          
\end{tabular}
\caption{Generation objectives for BART-RE, BART-DC, BART-DS and BART-DB on the TACRED dataset. This Table is the supplementary information for \S \ref{comparison_model}. Where BART-RE uses original relation names as the generation objectives. BART-DC, BART-DS and BART-DB drops correlation information, semantic information and both of them, respectively}
\label{app.tacred_data_v1}
\end{table*}

\begin{table*}[]
\centering
\begin{tabular}{l|l|l|l}
\hline
\textbf{BART-RE} & \textbf{BART-DC} & \textbf{BART-DS}         & \textbf{BART-DB}                                             \\ \hline
compare          & compare          & $\sim \sim \sim \sim$ & cffff                                                        \\
conjunction      & conjunction      & fff                      & ñ                                                            \\
evaluate for     & evaluate         & ö ---                    & \textless{}?                                                 \\
feature of       & feature          & -- cffffcc               & *=-                                                          \\
hyponym of       & hyponym of       & ()\{ cffffcc             & ö                                                            \\
part of          & part             & ē cffffcc                & \textgreater{}\textgreater{}\textbackslash{}\textbackslash{} \\
used for         & used for         & (): ---                  & ø                                                            \\ \hline
\end{tabular}
\caption{Generation objectives for BART-RE, BART-DC, BART-DS and BART-DB on the sciERC dataset. This Table is the supplementary information for \S \ref{comparison_model}.}
\label{app.scierc_data_v1}
\end{table*}

\begin{table*}[t]
\centering
\begin{tabular}{l|l|l}
\hline
\textbf{BART-Paraphrase}                                                     & \textbf{BART-Inquiry}                                                               & \textbf{BART-Synonym}                                                           \\ \hline
compare, comparison, like                                                    & compare, state art, relation between                                                & compare, contrast, comparison                                                   \\ \hline
conjunction, word, join                                                      & \begin{tabular}[c]{@{}l@{}}conjunction, lexicon driven, \\ followed by\end{tabular} & conjunction, link, joint                                                        \\ \hline
evaluate for, function, test                                                 & evaluate for, test set, error rate                                                  & \begin{tabular}[c]{@{}l@{}}evaluate for, \\ evaluate, assess\end{tabular}       \\ \hline
\begin{tabular}[c]{@{}l@{}}feature of, information, \\ describe\end{tabular} & feature of, tag, feature                                                            & \begin{tabular}[c]{@{}l@{}}feature of, typical, \\ characteristic,\end{tabular} \\ \hline
hyponym of, meaning, value                                                   & hyponym of, for example, such as                                                    & \begin{tabular}[c]{@{}l@{}}hyponym of, narrower, \\ subordinate\end{tabular}    \\ \hline
part of, section, part                                                       & part of, more details, background                                                   & part of, component, segment                                                     \\ \hline
used for, use, apply                                                         & used for, by using, can use                                                         & used for, apply to, employ                                                      \\ \hline
\end{tabular}
\caption{Generation objectives for BART-Paraphrase, BART-Inquiry and BART-Synonym on the sciERC dataset. This Table is the supplementary information for \S \ref{rela}. Different phrases are separated by comma.}
			\label{app.scierc_data_v2}
\end{table*}

\begin{table*}[]
\centering
\begin{tabular}{llll}
\hline
\textbf{BART-RE} & \textbf{BART-DC}  & \textbf{BART-DS}  & \textbf{BART-DB} \\ \hline
date of birth    & birthday is on    & ): cffffcc û      & cffff            \\
education degree & education degree  & 1111 +++          & ñ                \\
institution      & institution       & *=-               & \textless{}?     \\
place of birth   & was born in place & )+ cffffcc û      & *=-              \\
place of death   & place of death    & )+ cffffcc {]}{[} & ö               \\\hline
\end{tabular}
\caption{Generation objectives for BART-RE, BART-DC, BART-DS and BART-DB on the Google RE dataset. This Table is the supplementary information for \S \ref{comparison_model}.}
\label{app.googlere_data_v1}
\end{table*}

\begin{table*}[]
\begin{tabular}{l|l|l}
\hline
\textbf{BART-Paraphrase}                                                      & \textbf{BART-Inquiry}                                                                    & \textbf{BART-Synonym}                                                                  \\ \hline
date of birth, birth date, born                                               & \begin{tabular}[c]{@{}l@{}}date of birth, years old, \\ one founders\end{tabular}        & \begin{tabular}[c]{@{}l@{}}date of birth, birthday, \\ time of birth\end{tabular}      \\\hline
\begin{tabular}[c]{@{}l@{}}education degree, college,\\  student\end{tabular} & \begin{tabular}[c]{@{}l@{}}education degree, law school, \\ from university\end{tabular} & \begin{tabular}[c]{@{}l@{}}education degree, study, \\ graduate\end{tabular}           \\\hline
institution, group, community                                                 & institution, new york, click here                                                        & institution, organization, team                                                        \\\hline
place of birth, born, family                                                  & \begin{tabular}[c]{@{}l@{}}place of birth, first appearance, \\ first time\end{tabular}  & \begin{tabular}[c]{@{}l@{}}place of birth, birthplace, \\ born in a place\end{tabular} \\\hline
\begin{tabular}[c]{@{}l@{}}place of death, kill\\ death place\end{tabular}    & place of death, died on, who died                                                        & \begin{tabular}[c]{@{}l@{}}place of death, deathplace, \\ dead in a palce\end{tabular} \\\hline
\end{tabular}
\caption{Generation objectives for BART-Paraphrase, BART-Inquiry and BART-Synonym on the sciERC dataset. This Table is the supplementary information for \S \ref{rela}. Different phrases are separated by comma.}
\label{app.googlere_data_v2}
\end{table*}

\onecolumn
\begin{center}
\begin{longtable}{l|l|l}
\hline
\textbf{BART-Paraphrase} & \textbf{BART-Inquiry} & \textbf{BART-Synonym}                                        \\ \hline
\endhead
\begin{tabular}[c]{@{}l@{}}date of birth,\\ birth date, born\end{tabular}                               & \begin{tabular}[c]{@{}l@{}}date of birth,\\ henry bradford, \\ state resident\end{tabular}                   & \begin{tabular}[c]{@{}l@{}}date of birth, birthday, \\ time of birth\end{tabular}                                                 \\ \hline
\begin{tabular}[c]{@{}l@{}}top member employee,\\ top level, ceo\end{tabular}                           & \begin{tabular}[c]{@{}l@{}}top member employee,\\ mf global, united states\end{tabular}                      & \begin{tabular}[c]{@{}l@{}}top member employee, \\ chairman, president\end{tabular}                                               \\ \hline
no relation                                                                                             & no relation                                                                                                  & no relation                                                                                                                       \\ \hline
\begin{tabular}[c]{@{}l@{}}city of residence,\\ live, urban\end{tabular}                                & \begin{tabular}[c]{@{}l@{}}city of residence,\\ los angeles, east berlin\end{tabular}                        & \begin{tabular}[c]{@{}l@{}}city of residence, \\ live in city, hometown\end{tabular}                                              \\ \hline
\begin{tabular}[c]{@{}l@{}}person parent,\\ parent, mom\end{tabular}                                    & \begin{tabular}[c]{@{}l@{}}person parent, \\ philip mayer, kaiser\end{tabular}                               & \begin{tabular}[c]{@{}l@{}}person parent, \\ father, mother\end{tabular}                                                          \\ \hline
\begin{tabular}[c]{@{}l@{}}country of headquarters\\ capital, government\end{tabular}                   & \begin{tabular}[c]{@{}l@{}}country of headquarters,\\ hong kong, ghent university,\end{tabular}              & \begin{tabular}[c]{@{}l@{}}country of headquarters, \\ capital, city\end{tabular}                                                 \\ \hline
\begin{tabular}[c]{@{}l@{}}state or provinces of residence,\\ citizenship, residence\end{tabular}       & \begin{tabular}[c]{@{}l@{}}state or provinces of residence,\\ paul gillmor, millender mcdonald\end{tabular}  & \begin{tabular}[c]{@{}l@{}}state or provinces of \\ residence, hometown\\ live in state province,\end{tabular}                    \\ \hline
\begin{tabular}[c]{@{}l@{}}other family,\\ family, relatives\end{tabular}                               & \begin{tabular}[c]{@{}l@{}}other family, saddam kamel, \\ two sons\end{tabular}                              & \begin{tabular}[c]{@{}l@{}}other family, \\ uncle, aunt\end{tabular}                                                              \\ \hline
\begin{tabular}[c]{@{}l@{}}city of birth,\\ birth city, born\end{tabular}                               & \begin{tabular}[c]{@{}l@{}}city of birth,\\ grew up, born on\end{tabular}                                    & \begin{tabular}[c]{@{}l@{}}city of birth, born in city,\\  birth place\end{tabular}                                               \\ \hline
religion, belief, islam                                                                                 & \begin{tabular}[c]{@{}l@{}}religion, jewish ness, \\ catholic church\end{tabular}                            & \begin{tabular}[c]{@{}l@{}}religion, belongs to, \\ believe in\end{tabular}                                                       \\ \hline
members,group, partner                                                                                  & \begin{tabular}[c]{@{}l@{}}members, oic membership, \\ coca cola\end{tabular}                                & members, static,affiliates                                                                                                        \\ \hline
spouse, wife, husband                                                                                   & spouse, doing with, married                                                                                  & spouse, couple, wife                                                                                                              \\ \hline
\begin{tabular}[c]{@{}l@{}}number of employees member,\\ employee number, sum\end{tabular}              & \begin{tabular}[c]{@{}l@{}}number of employees member,\\ nuclear suppliers, \\ central american\end{tabular} & \begin{tabular}[c]{@{}l@{}}number of employees member, \\ total employees, employees\end{tabular}                                 \\ \hline
origin, source, root                                                                                    & \begin{tabular}[c]{@{}l@{}}origin,\\ african american, \\ prime minister\end{tabular}                        & \begin{tabular}[c]{@{}l@{}}origin, nationality, \\ background\end{tabular}                                                        \\ \hline
\begin{tabular}[c]{@{}l@{}}title, character, \\ name description\end{tabular}                           & \begin{tabular}[c]{@{}l@{}}title, don know,\\ no business\end{tabular}                                       & title, is a, honor                                                                                                                \\ \hline
\begin{tabular}[c]{@{}l@{}}political religious affiliation,\\ church, believe in\end{tabular}           & \begin{tabular}[c]{@{}l@{}}political religious affiliation,\\ liberation front, moro national\end{tabular}   & \begin{tabular}[c]{@{}l@{}}political religious affiliation, \\ religion, belief\end{tabular}                                      \\ \hline
\begin{tabular}[c]{@{}l@{}}state or province of death,\\ state, dead\end{tabular}                       & \begin{tabular}[c]{@{}l@{}}state or province of death,\\ joseph meskill, henry sheldon\end{tabular}          & \begin{tabular}[c]{@{}l@{}}state or province of death, \\ died in state or province,\\ death place\end{tabular}                   \\ \hline
\begin{tabular}[c]{@{}l@{}}school attended,\\ student, college\end{tabular}                             & \begin{tabular}[c]{@{}l@{}}school attended,\\ law school, state university\end{tabular}                      & \begin{tabular}[c]{@{}l@{}}school attended, studied in, \\ graduated from\end{tabular}                                            \\ \hline
\begin{tabular}[c]{@{}l@{}}cause of death,\\ die, killed\end{tabular}                                   & \begin{tabular}[c]{@{}l@{}}cause of death,\\ heart failure, so ill\end{tabular}                              & \begin{tabular}[c]{@{}l@{}}cause of death, reason of death, \\ death reason\end{tabular}                                          \\ \hline
\begin{tabular}[c]{@{}l@{}}country of birth,\\ born, birthplace\end{tabular}                            & \begin{tabular}[c]{@{}l@{}}country of birth,\\ boyle scotland, born glasgow\end{tabular}                     & \begin{tabular}[c]{@{}l@{}}country of birth, \\ born in country, birth place\end{tabular}                                         \\ \hline
\begin{tabular}[c]{@{}l@{}}siblings, brother,\\ family members,\end{tabular}                            & \begin{tabular}[c]{@{}l@{}}siblings, martin luther, \\ impressed with\end{tabular}                           & siblings, brother, sister                                                                                                         \\ \hline
employee of, work, hired                                                                                & \begin{tabular}[c]{@{}l@{}}employee of,\\ white house, \\ involved business\end{tabular}                     & employee of, staff of, hire                                                                                                       \\ \hline
\begin{tabular}[c]{@{}l@{}}city of headquarters,\\ district, city\end{tabular}                          & \begin{tabular}[c]{@{}l@{}}city of headquarters,\\ highland capital, tool works\end{tabular}                 & \begin{tabular}[c]{@{}l@{}}city of headquarters, \\ located in city, central city\end{tabular}                                    \\ \hline
children, child, kid                                                                                    & \begin{tabular}[c]{@{}l@{}}children,\\ two sons, years old\end{tabular}                                      & children, son, daughter                                                                                                           \\ \hline
\begin{tabular}[c]{@{}l@{}}country of death,\\ death of, disease\end{tabular}                           & \begin{tabular}[c]{@{}l@{}}country of death, \\ lebanon, caucus\end{tabular}                                 & \begin{tabular}[c]{@{}l@{}}country of death, death place,\\ died in a country\end{tabular}                                        \\ \hline
\begin{tabular}[c]{@{}l@{}}shareholder,\\ owner, person to hold\end{tabular}                            & \begin{tabular}[c]{@{}l@{}}shareholder,\\ yield, general atlantic\end{tabular}                               & \begin{tabular}[c]{@{}l@{}}shareholder, stockholder,\\  holds shares in\end{tabular}                                              \\ \hline
\begin{tabular}[c]{@{}l@{}}member of,\\ have power, member of group\end{tabular}                        & \begin{tabular}[c]{@{}l@{}}member of, red sox, \\ universal studios\end{tabular}                             & \begin{tabular}[c]{@{}l@{}}member of, member, \\ member management\end{tabular}                                                   \\ \hline
\begin{tabular}[c]{@{}l@{}}organization parent,\\ hierarchy, parent organize\end{tabular}               & \begin{tabular}[c]{@{}l@{}}organization parent,\\ carnival cruise, \\ space center\end{tabular}              & \begin{tabular}[c]{@{}l@{}}organization parent, \\ parent company, parent firm\end{tabular}                                       \\ \hline
website, url, http                                                                                      & \begin{tabular}[c]{@{}l@{}}website, http www, \\ facebook com\end{tabular}                                   & website, url, site                                                                                                                \\ \hline
age, year, time                                                                                         & \begin{tabular}[c]{@{}l@{}}age, year old, \\ relationship between\end{tabular}                               & age, year                                                                                                                         \\ \hline
\begin{tabular}[c]{@{}l@{}}dissolved,\\ expulsion, dissolution\end{tabular}                             & \begin{tabular}[c]{@{}l@{}}dissolved, liberty media, \\ american basketball\end{tabular}                     & dissolved, relieve, disband                                                                                                       \\ \hline
\begin{tabular}[c]{@{}l@{}}state or province of birth,\\ state, born\end{tabular}                       & \begin{tabular}[c]{@{}l@{}}state or province of birth,\\ harry dent, brigham young\end{tabular}              & \begin{tabular}[c]{@{}l@{}}state or province of birth, \\ born in state province, \\ birth place\end{tabular}                     \\ \hline
\begin{tabular}[c]{@{}l@{}}subsidiary,\\ subdivision, corporation\end{tabular}                          & \begin{tabular}[c]{@{}l@{}}subsidiary, carnival, \\ cruise lines\end{tabular}                                & subsidiary, branch, affiliate                                                                                                     \\ \hline
\begin{tabular}[c]{@{}l@{}}date of death,\\ death date, period of\end{tabular}                          & \begin{tabular}[c]{@{}l@{}}date of death,\\ who had, first black\end{tabular}                                & date of death, died time, death                                                                                                   \\ \hline
\begin{tabular}[c]{@{}l@{}}founded by,\\ created by, by origin\end{tabular}                             & \begin{tabular}[c]{@{}l@{}}founded by,\\ focus on family, \\ jerusalem foundation\end{tabular}               & founded by, founder, originator                                                                                                   \\ \hline
\begin{tabular}[c]{@{}l@{}}country of residence,\\ town, citizen\end{tabular}                           & \begin{tabular}[c]{@{}l@{}}country of residence,\\ cannot here, there no\end{tabular}                        & \begin{tabular}[c]{@{}l@{}}country of residence, \\ live in country, motherland\end{tabular}                                      \\ \hline
\begin{tabular}[c]{@{}l@{}}person alternate name,\\ person alternate, \\ name of alternate\end{tabular} & \begin{tabular}[c]{@{}l@{}}person alternate name,\\ real name, good friends\end{tabular}                     & \begin{tabular}[c]{@{}l@{}}person alternate name,\\ also known as, people alias\end{tabular}                                      \\ \hline
\begin{tabular}[c]{@{}l@{}}city of death, hell,\\ city of death in,\end{tabular}                        & \begin{tabular}[c]{@{}l@{}}city of death,\\ prime minister, grand canyon\end{tabular}                        & \begin{tabular}[c]{@{}l@{}}city of death, died in city, \\ death place\end{tabular}                                               \\ \hline
\begin{tabular}[c]{@{}l@{}}state or province of headquarters,\\ state, headquarters\end{tabular}        & \begin{tabular}[c]{@{}l@{}}state or province of headquarters,\\ kennedy space, space center\end{tabular}     & \begin{tabular}[c]{@{}l@{}}state or province of \\ headquarters, \\ located in state province, \\ provincial capital\end{tabular} \\ \hline
\begin{tabular}[c]{@{}l@{}}organization alternate name,\\ group title, name of\end{tabular}             & \begin{tabular}[c]{@{}l@{}}organization alternate name,\\ liberation front, \\ moro national\end{tabular}    & \begin{tabular}[c]{@{}l@{}}organization alternate name, \\ also known as, \\ organization alias\end{tabular}                      \\ \hline
\begin{tabular}[c]{@{}l@{}}charges, function, \\ element in list\end{tabular}                           & \begin{tabular}[c]{@{}l@{}}charges,charged with, \\ second degree\end{tabular}                               & charges, convict, accuse                                                                                                          \\ \hline
\begin{tabular}[c]{@{}l@{}}founded, author, \\ established\end{tabular}                                 & \begin{tabular}[c]{@{}l@{}}founded, theme park,\\ american parliament\end{tabular}                           & founded, establish, build                                                                                                         \\ \hline
\caption{Generation objectives for BART-Paraphrase, BART-Inquiry and BART-Synonym on the TACRED dataset. This Table is the supplementary information for \S \ref{rela}. Different phrases are separated by comma.}
			\label{app.tacred_data_v2}\\
\end{longtable}
\end{center}
\twocolumn

\end{document}